%% file: main.tex
\documentclass[journal]{IEEEtran} 

\IEEEoverridecommandlockouts 
\usepackage{paralist}
\usepackage{url}
\usepackage{caption}
\usepackage{amsmath, amssymb, amssymb, amsfonts}
\usepackage{algorithm}
\usepackage{algorithmicx}
\usepackage{algpseudocode}
\usepackage{leftidx}
\usepackage{bm, bbm}
\usepackage{mathrsfs}
\usepackage{textcomp}
\usepackage{graphicx, wrapfig, lipsum}
\usepackage[hidelinks]{hyperref}
\usepackage{mathtools}
\usepackage{multirow}
\usepackage{subcaption}
\usepackage{makecell}
\usepackage{tabularx}
\usepackage[noadjust]{cite}
\usepackage{array}
\usepackage{stfloats}
\usepackage{verbatim}
\usepackage{balance}
\usepackage{tabularray}
\usepackage{lettrine}
\usepackage{xcolor}
\usepackage{tikz}
\usetikzlibrary{arrows.meta}
\usepackage[all]{comment} 

\usepackage{soul}

\newtheorem{problem}{Problem}
\DeclareMathOperator{\atantwo}{atan2}

\newcommand{\real}[1][ ]{\mathbb{R}^#1}
\newcommand{\norm}[1]{\lVert #1 \rVert}

\newcommand{\vect}[1]{\mathbf{#1}}
\newcommand{\mat}[1]{\mathbf{#1}}

\newcommand{\diff}[3]{\mathbf{#1}_#2 -\mathbf{#1}_#3}
\newcommand{\neigh}[1]{\mathcal{N}_#1}

\newcommand{\overbar}[1]{\mkern 1.5mu\overline{\mkern-1.5mu#1\mkern-1.5mu} \mkern
1.5mu}

\begin{document}
        \title{On Swarm Leader Identification using Probing Policies}

        \author{Stergios E. Bachoumas, \textit{IEEE Member}
        and~Panagiotis~Artemiadis$^*$, \textit{IEEE Senior Member}

        \thanks{Stergios E. Bachoumas and Panagiotis Artemiadis are with the Mechanical Engineering Department, at the University of Delaware, Newark, DE 19716, USA. {\tt\small stevbach@udel.edu, partem@udel.edu}}
        \thanks{$^*$Corrresponding author}}%

        \maketitle

        \begin{abstract}
            Identifying the leader within a robotic swarm is crucial, especially in adversarial contexts where leader concealment is necessary for mission success. This work introduces the interactive Swarm Leader Identification (iSLI) problem, a novel approach where an adversarial probing agent identifies a swarm's leader by physically interacting with its members. We formulate the iSLI problem as a Partially Observable Markov Decision Process (POMDP)  and employ Deep Reinforcement Learning, specifically Proximal Policy Optimization (PPO), to train the prober's policy. The proposed approach utilizes a novel neural network architecture featuring a Timed Graph Relationformer (TGR) layer combined with a Simplified Structured State Space Sequence (S5) model. The TGR layer effectively processes graph-based observations of the swarm, capturing temporal dependencies and fusing relational information using a learned gating mechanism to generate informative representations for policy learning. Extensive simulations demonstrate that our TGR-based model outperforms baseline graph neural network architectures and exhibits significant zero-shot generalization capabilities across varying swarm sizes and speeds different from those used during training. The trained prober achieves high accuracy in identifying the leader, maintaining performance even in out-of-training distribution scenarios, and showing appropriate confidence levels in its predictions. Real-world experiments with physical robots further validate the approach, confirming successful sim-to-real transfer and robustness to dynamic changes, such as unexpected agent disconnections.
        \end{abstract}

        \textbf{{\textit{Note to Practitioners}} - This paper provides a framework that can be directly applied to enhance the security and resilience of multi-agent robotic systems, particularly in scenarios where maintaining operational secrecy is critical. For practitioners in fields such as defense, autonomous surveillance, and logistics, the methods presented here offer a novel approach to stress-testing and improving the robustness of their robotic swarms. The core contribution, an intelligent ``prober'' agent trained via deep reinforcement learning, can be adapted as a versatile tool for systematically identifying and mitigating vulnerabilities in leader-follower systems prior to deployment. By using our framework, engineers can implement this adversarial training methodology to develop swarms that are more resilient to intelligent attacks, ultimately leading to more secure and reliable real-world applications of autonomous systems.}
                
        Video: \href{https://youtu.be/sTQK14R3gtM}{https://youtu.be/sTQK14R3gtM}\\\\

        \input{intro}
        \input{prob_form}

        \input{env}
        \input{policy}
        \input{sim_experiments}
        \input{sim_results}

        \input{real_experiments}
        \input{conclusions}

        \balance
        \bibliographystyle{bibliography/IEEEtran}
        \bibliography{bibliography/references}
\end{document}

%% file: intro.tex
\section{Introduction}

\lettrine{I}{n} nature, the utilization of collective behaviors (i.e., swarming) is abundant. Biological entities are known to group with their peers to defend themselves, conceal their goals, and engage in social interaction and information sharing. As biological agents continue to inspire robotic teams, the use of swarm robotics is becoming increasingly prevalent. The advantages of swarm robotic systems primarily derive from their scalability and the relatively low cost of individual units~\cite{kolling2015human}. However, these same characteristics introduce significant challenges in Human-Swarm Interaction (HSI) contexts, particularly regarding an operator's ability to effectively control the swarm while maintaining adequate situational awareness to accomplish mission objectives.

Swarms inherently manifest complex emergent behaviors at the system level, such as flocking patterns, which human operators often struggle to interpret accurately. Previous research has demonstrated that robotic mission failures frequently stem from diminished situational awareness when operators experience cognitive overload~\cite{liu2013robotic}. Among the various control methodologies proposed in the literature, the leader-follower paradigm has emerged as particularly effective for swarm control~\cite{kolling2015human ,Deka1, deka2020naturalemergenceheterogeneousstrategies,desai,leader_flocking}. In this paradigm, a designated leader agent has knowledge of the task and guides the swarm toward a common goal, while the remaining agents, who do not have knowledge of the task, follow the leader's actions. This approach simplifies the control problem by reducing the number of agents that the operator must communicate with or even control directly, thereby enhancing the operator's situational awareness and reducing cognitive load.

In contested environments, identifying and neutralizing the leader can effectively decapitate the swarm, crippling its operational capacity \cite{Deka1,Zheng_2020}. Thus, the leader of a robotic swarm can be its Achilles' heel, a critical vulnerability that an intelligent adversary will seek to exploit. While prior research has explored the problem of leader identification in swarms, it has largely focused on passive observation-based solutions, where the leader is identified by observing the motion of the swarm. For instance, in ~\cite{Ajitesh}, the authors developed a probabilistic approach based on DBSCAN clustering to identify swarm leaders. Similarly, in ~\cite{Zheng_2020}, a Genetic Algorithm (GA) was proposed to optimize private flocking mechanisms, wherein the swarm conceals its leader from an adversarial discriminator. Additionally, the authors in ~\cite{Deka1} implemented a Graph Neural Network (GNN) framework to model leader concealment strategies against an artificial adversary represented by a Long Short-Term Memory (LSTM) neural network. While these studies collectively demonstrate the growing interest in computational approaches to the SLI problem, they do not explore the interactive-based scenario that can arise in real-world applications.

\begin{figure}[t!]
    \centering
    \includegraphics[width=1\columnwidth]{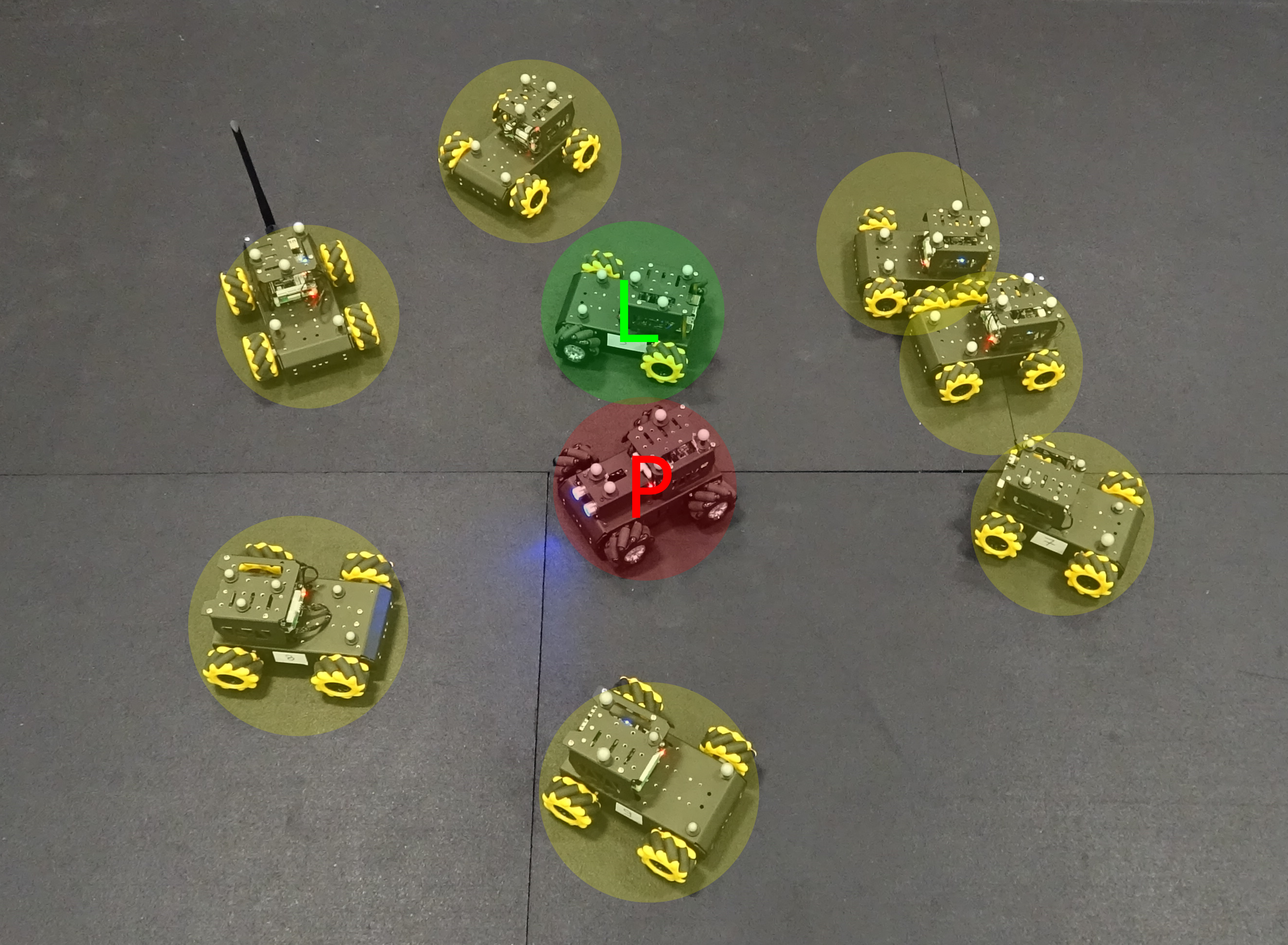}
    \vspace{-0.2in}
    \caption{A flocking swarm of robots (yellow circles) and their leader (green circle), move toward a common goal. An adversary-prober
    (red circle) strategically maneuvers within the swarm ranks in order to identify the leader agent.}
        \vspace{-0.2in}
    \label{fig:intro}
\end{figure}

We propose that a more effective adversary is one that abandons passive observation and instead seeks to provoke the swarm, as analyzing the swarm's reactions to targeted interactions is the key to unmasking its chain of command. In a recent paper \cite{bachoumas_icra_2025}, the authors introduced the interactive Swarm Leader Identification (iSLI) problem and tackled a simplified version of it. However, their assumptions of a fixed number of agents and a constant world size imposed a hard constraint on the agent distances, making their policy unable to generalize to new environments.

Within the broader context of Multi-Agent Reinforcement Learning (MARL), the authors in \cite{HAMA} investigate multi-agent cooperation by modeling agent interactions with GNNs. Although this approach yields valuable insights for predator-prey scenarios, its applicability is limited by the simplicity of the evaluated cases. The complexity of these scenarios does not compare to that of the iSLI problem, which scales significantly with the number of agents in the swarm.

To address the limitations of prior research on the iSLI and similar problems, this work investigates the scenario in which an adversary attempts to identify a swarm's hidden leader within a partially observable environment, as depicted in Fig. \ref{fig:intro}. To identify the leader, the adversary must interact with the swarm and use its interactions as a guide to solve the problem. Building on \cite{bachoumas_icra_2025}, we model the iSLI problem using the Partially Observable Markov Decision Process (POMDP) framework \cite{astrom_pomdp} and leverage Proximal Policy Optimization 
(PPO) \cite{PPO,huang2022cleanrl}, a Deep Reinforcement Learning (DRL) \cite{DRL,mnih2015humanlevel} algorithm, to train the adversarial prober's policy, guiding it toward identifying the leader agent within the swarm.  Contrary to \cite{bachoumas_icra_2025}, inspired by \cite{HAMA, agarwal2019learningtransferablecooperativebehavior, deka2020naturalemergenceheterogeneousstrategies}, we introduce a novel GNN layer that leverages the interactions of the prober with the swarm members in order to effectively modulate the information in the graph. This fundamental distinction from the approach in \cite{bachoumas_icra_2025} renders our policy both permutation invariant and effective in environments with a variable number of swarm agents. 

Additionally, motivated by the strong performance of modern structure state space sequence models \cite{S4,smith2022simplified} in long sequence learning tasks, we utilize a Simplified Structured State Space Sequence (S5) encoder \cite{smith2022simplified} and benefit from its strong generalization capabilities in POMDPs as demonstated from earlier works in DRL \cite{bachoumas_icra_2025,S5RL,jeen2025zeroshotreinforcementlearningpartial}. We validate our method by demonstrating that the trained prober accurately identifies the leader with high confidence, even in out-of-distribution scenarios in simulated environments. Moreover, real-world experiments with physical robots validate the successful transfer of the policy from simulation to reality and confirm its robustness against dynamic changes, such as unexpected agent disconnections.

The contributions of this work can be summarized as follows:
\begin{itemize}
    \item A novel POMDP formulation of the iSLI environment, that leverages graphs to enable and accelerates future research in adversarial swarm robotics across different environment and swarm sizes.
    \item A novel Graph Neural Network (GNN) layer that uses interaction-based relations as a neural gating mechanism \cite{lstm} to effectively solve the iSLI problem in varying scenarios.
    \item A training methodology for an adversarial prober that generalizes across diverse environments and swarm sizes, demonstrating efficacy in both simulated and real-world applications.
\end{itemize}

This work makes a significant contribution to the robotics community by proposing a novel paradigm for achieving swarm resilience. Instead of focusing solely on defensive strategies, we believe that the development of truly resilient swarms necessitates the creation of intelligent adversaries. By first engineering an intelligent \textit{prober} agent adept at identifying a swarm's leader through active engagement, we establish a powerful tool for adversarial training. This prober can then be utilized to compel the leader-based swarm to evolve its maneuvers into more sophisticated and decentralized-appearing behaviors, effectively forging resilience by forcing the system to train against its own worst-case vulnerability. This \textit{offense-as-defense} methodology provides a new framework for systematically exposing and mitigating weaknesses in swarm architectures, thereby advancing the development of robust and secure multi-agent systems.

%% file: prob_form.tex
\section{Problem Formulation}

\subsection{Swarm Graph}

In this article, the swarm is modeled as a dynamic, directed topological graph $\mathcal{G}_t=(\mathcal{H}_t,\mathcal{E}_t)$~\cite{gnn_review_2021,ZHAO2024109166}, where $\mathcal{H}_t$ is the time-varying node set, and $\mathcal{E}_t$ the set of communication links.

\subsubsection{Graph Nodes}

The spatial configuration of the nodes determines the topology of the graph. With each node, we associate a feature vector representation, denoted by $\vect{h} _{i}\in \real{D_n}$ for $i=1,\ldots,N$, and $D_n$ the dimensionality of the node feature vectors. Thus, the node set is defined as the set of all node feature vectors $\mathcal{H}(t) \coloneqq \{\vect{h}_{1}(t),\cdots \vect{h}_{N} (t)\}$ at time $t$. Note that, as this is a set, ideally permutation-invariant functions, i.e., functions whose output does not change when the order of the nodes (agents) changes, should be used to handle it.

\subsubsection{Graph Edges}

In the graph, agents $i$ and $j$ are connected if the distance $d_{ij}$ between them is less than a predefined communication radius $d$, i.e., $d_{ij}(t) \leq d$ \cite{adv_resilience} with $d$ being the same for all agents. 

The structure of $\mathcal{G}$ is inherently asymmetric to reflect the leader's unique role and increased influence. We formalize this by assigning a scalar weight to each edge based on the identity of the outgoing agent (sender). Edges originating from the leader are assigned a weight $w_L$, while those from follower agents are assigned a weight $w_F$. To ensure that the leader's influence is dominant across the network, we impose the condition $w_L>w_F$. This weighting scheme is the primary mechanism that distinguishes the leader from the followers within the swarm's communication topology.

With each existing directed edge, we associate a feature vector representation, denoted by $\vect{e}_{i\rightarrow j}(t) \in \real {D_e}$ that is the directed edge from agent $i$ to $j$ ($i,j=1,\ldots,N$), where $D_e$ is the dimensionality of the edge feature vector. For convenience, all nodes have self-edges, i.e., edges that loop back to themselves. The edge set is thus defined as $\mathcal{E}_{t}\coloneqq \{\vect{e}_{i\rightarrow j}(t)|\forall i,j=0,\cdots,N\}$ and stores all the edges in the graph.

\subsection{Problem Definition}

The iSLI problem presents a significant challenge from the prober's perspective due to two key factors: \textbf{partial observability of the environment and unknown swarm dynamics}. The prober's limited access to complete spatial and temporal information, coupled with its inability to predict the swarm's motion, necessitates a sequential decision-making framework. Consequently, the prober must leverage a history of observations, as past information may remain critical for making optimal decisions in the present.

Specifically, using the definitions we just made, the swarm graph $\mathcal{G}_t$ and the information encoded in the node set $\mathcal{H}_t$ and the edge set $\mathcal{E}_t$ are not fully observed by the prober. For instance, the prober does not know the underlying connectivity of the graph and thus does not know the hidden relations between the agents of the swarm. Moreover, the prober does not observe the swarm continuously, but uses its sensors at a fixed frequency. Thus, we model the information available to the prober at each time step $k$ using a graph snapshot $\hat{\mathcal{G}}[k]=(\hat{\mathcal{H}}[k],\hat{\mathcal{E}}[k], k)$. We defined the specifics of the observed node set $\hat{\mathcal{H}}[k]$ and edge set $\hat{\mathcal{E}}[k]$ in the next section.

The primary problem investigated in this study is formalized as follows:

\begin{problem}
    [Interactive Swarm Leader Identification] Given observations of a swarm $\mathcal{S}$ performing a flocking task $\mathcal{T}_{s}$ in an obstacle-free space $\mathcal{F}$, the prober $p$ is tasked with identifying the leader $L$, i.e., the most influential agent (node) by forcefully interacting with the agents of the swarm in as many as $\bar{K}$ time steps. The prober is assumed to have no prior knowledge of the underlying swarm dynamics. At any given timestep $k$, where $0 \le k \le \bar{K}$, its available information is limited to the sequence of partial graph snapshots observed up to that point: $\mathcal{\hat{G}}_{0:k} \coloneqq [\mathcal{\hat{G}}_{0}, \mathcal{\hat{G}}_{1}, \cdots, \mathcal{\hat{G}}_{k}]$.
\end{problem}

%% file: env.tex
\section{The iSLI Environment}

In this work, we build on the environment developed in \cite{bachoumas_icra_2025}, which used double integrator dynamics for the swarm agents, introducing significant modifications to better suit our objectives. This is done because of two key limitations that we identified with using the double integrator dynamics:

\subsubsection{Absence of orientation} The absence of orientation, for the agents, significantly limits the richness of flocking behaviors of the swarm and is directly related to the difficulty of the iSLI task.

\subsubsection{Sim-to-real gap} The difference between simulated and real dynamics makes the reliable transfer of learned policies trained in simulation to reality challenging without further training.

\vspace{-8pt}
\subsection{Agent Dynamics}
To tackle the aforementioned limitations, in this work, the prober and the $N$ members of the swarm each follow overdamped boid-like dynamics inspired by \cite{Reynolds}:

\begin{align}
    \label{eq:flock_dynamics}
    \dot{\vect{p}}_{i}(t)  & = -\nabla_{\vect{p}_i}E\big(\mat{P}(t),\Theta(t)\big) + v_{max}\vect{\hat{n}}_{i}(t) +\vect{f}_{pi}(t)
    \\
    \dot{\theta}_i(t) &= -\nabla_{\theta_i}E\big(\mat{P}(t),\Theta(t)\big)
    \nonumber \\
    \dot{\vect{p}}_{p}(t) & = \vect{v}_{p}(t) -\sum_{i=1}^N\vect{f}_{pi}(t) \\
    \theta_p(t) &= \atantwo\big(\dot{y}_{p},\dot{x}_{p}\big) \nonumber
\end{align}

Let $E$ be an energy function that the agents of the swarm seek to minimize with their movement and $\vect{p}_{i}(t) = [x_{i}, y_{i}]^{\top} \in \mathbb{R}^{2}$, $\vect{n}_{i}(t) = [\cos(\theta_{i}), \sin(\theta_{i})]^{\top} \in [-1,1]^{2}$ and $\theta_{i}(t) \in [-\pi, \pi]$ for $i=1, 2, \ldots, N$ be the position vector, direction vector and orientation of the $i$\textsuperscript{th}
swarm agent, respectively. Likewise, let
$\vect{p}_{p}(t) = [x_{p}, y_{p}]^{\top} \in \mathbb{R}^{2}$, $\vect{v}_{p}(t) =
[\dot{x}_{p}, \dot{y}_{p}]^{\top} \in \mathbb{R}^{2}$ and $\theta_{p} \in [-\pi,\pi]$ be the position vector, velocity vector and orientation of the prober, respectively. The orientation of the prober is defined as the heading angle based on its velocity vector components, i.e., $\theta_p(t) =  \atantwo\big(\dot{y}_{p},\dot{x}_{p}\big)$.
Time is denoted by
$t\in \real{}_{\ge 0}$.  Additional forces $\vect{f}_{pi}(t)$ model the interactions between the prober and the swarm that act as velocity terms in the dynamics. For convenience of notation, we collect the positions of all swarm agents in an array:
$\mat{P}(t) \coloneqq [\vect{p}_{1}, \dots , \vect{p}_{N}]^{\top} \in
\mathbb{R}^{N\times2}$
and all orientations in vector $\Theta(t) \coloneqq [\theta_{1}, \dots, \theta_{N}]^{\top} \in [0,\pi)^{N}$. 
We note that all physics-related quantities use the SI system of units.

\subsection{Swarm Flocking Behavior}
The swarm's emergent flocking behavior is generated by an energy-based model founded on the three well-known principles of Reynolds \cite{Reynolds}: alignment, cohesion, and separation. From this model, we derive the forces exerted on each agent, as specified in (\ref{eq:flock_dynamics}). The total energy function is defined as follows:
\begin{equation}
    E(\mat{P},\Theta)=E_{align}(\mat{P},\Theta)+E_{separate}(\mat{P})+E_{cohere}(\mat{P})
\end{equation}
where each term is defined below.

\subsubsection{Cohesion}
For each of the followers $i$, we define its neighborhood $\mathcal{N}_i$, as all other agents $j$ that are one hop distance away in the swarm graph. Topologically, this is equivalent to $\norm{\Delta \vect{p}_{ij}}=d_{ij} < d_{coh}$, where $\Delta \vect{p}_{ij} = \vect{p}_i - \vect{p}_j$ is the relative position vector between agents $i$ and $j$, and $\norm{\Delta \vect{p}_{ij}}$ its Euclidean norm. The leader agent, $L$, exerts a distinct, global influence, weighted by $w_L$, on all follower agents irrespective of their proximity. Accordingly, for each agent $i$ we define its displacement vector $\Delta \vect{p}_i$ based on its relative position from its one-hop neighbors $\mathcal{N}_i$ and the leader $L$ as:
\begin{equation}
 \Delta \vect{p}_i = \frac{1}{|\mathcal{N}_i|+w_L} \Bigg(\sum_{j \in \mathcal{N}_i} \Delta \vect{p}_{ij} + w_L\Delta \vect{p}_{iL}\Bigg)
 \label{eq:com_disp}
\end{equation}
We then define the unit direction vector $\hat{\Delta \vect{p}}_i$ as follows: $\hat{\Delta \vect{p}}_i = \Delta \vect{p}_i / \norm{\Delta \vect{p}_i}$. The cohesion energy contribution for agent $i$ aims to align its heading vector, $\hat{\vect{n}}_i$, with this direction, and is given by:

\begin{equation}
 E_{cohere} = \frac{1}{2} w_{coh} (1 - \hat{\Delta \vect{p}}_i \cdot \hat{\vect{n}}_i)^2
 \label{eq:cohesion_energy}
\end{equation}
where the weight parameter $w_{coh}$ determines the strength of the cohesion force. This energy is minimized when the agent $i$ is pointing directly towards the midpoint between the centroid of its neighbors and the leader's position (i.e., when $\hat{\Delta \vect{p}}_i \cdot \vect{n}_i = 1$).  Crucially, when calculating the force derived from this energy (typically via the negative gradient operator $-\nabla()$), a stop-gradient is applied to the vector $\Delta \vect{p}_i$. This formulation ensures that the cohesion rule only governs the agent's orientation, i.e., its turning towards the group center, rather than independently inducing translational displacement.
\subsubsection{Alignment}
For a pair of agents $i$ and $j$, the alignment interaction energy is non-zero only if they are within a distance $d_{al}$ of each other ($\norm{\Delta \vect{p}_{ij}} < d_{al}$). The energy function is designed to be minimized when their heading vectors, $\hat{\vect{n}}_i$ and $\hat{\vect{n}}_j$, are parallel ($\hat{\vect{n}}_i \cdot \hat{\vect{n}}_j = 1$) and maximized when they are anti-parallel ($\hat{\vect{n}}_i \cdot \hat{\vect{n}}_j = -1$). The energy also increases smoothly as the agents get closer within the interaction radius. Thus, the alignment energy is given by:

\begin{equation}
E_{align} =
\begin{cases}
\frac{w_{al}}{\alpha} \left(1 - \frac{\norm{\Delta \vect{p}_{ij}}}{d_{al}}\right)^\alpha (1 - \hat{\vect{n}}_i \cdot \hat{\vect{n}}_j)^2 & \text{, if } \frac{\norm{\Delta \vect{p}_{ij}}}{d_{al}} < 1 \\
0 & \text{, otherwise}
\end{cases}
\label{eq:align_energy}
\end{equation}

Here, $w_{al}$ controls the strength of the alignment tendency, and $\alpha$ is a parameter influencing the shape of the potential well related to distance. As with cohesion, the primary goal of alignment is to adjust orientation. Therefore, when calculating forces, a stop-gradient is applied to the term $\Delta \vect{p}_{ij}$ to ensure that agents primarily \textit{turn} to align with neighbors, rather than being pushed apart positionally by this energy term.

\subsubsection{Separation}

This is modeled by considering a repulsive potential that activates only when the distance between two agents $i$ and $j$, $\norm{\Delta \vect{p}_{ij}}$, falls below a specific threshold $d_{sep}$. A simple energy function for separation is:

\begin{equation}
 E_{separate}=
 \begin{cases}
  \frac{w_{sep}}{\alpha} \left(1 - \tfrac{\norm{\Delta \vect{p}_{ij}}}{d_{sep}}\right)^\alpha & \text{, if } \frac{\norm{\Delta \vect{p}_{ij}}}{d_{sep}} < 1 \\
  0 & \text{, otherwise}
 \end{cases}
 \label{eq:avoid_energy}
\end{equation}

The weight $w_{sep}$ determines the strength of the repulsion. This energy increases sharply as agents approach each other within the $d_{sep}$ radius. 

All the agents of the swarm move to a target position by following the interaction from the leader. The leader is controlled to reach a desired position in space using the following velocity control law:
\vspace{-10pt}
\begin{equation}
    \begin{aligned}
        \vect{v}_L(\mat{P},\vect{p}_{G}) &= K_{p}(\vect{p}_{G}-\vect{p}_{L}) + K_{i} \int_{0}^{t} (\vect{p}_{G}-\vect{p}_L(\tau))d\tau \\
        &\quad + K_{L\alpha}(\bar{\vect{p}}_{\alpha}-\vect{p}_{L})
    \end{aligned}
\end{equation}
where $K_{p}$, $K_i$, and $K_{L\alpha}\in \real{}$ are the gains of the leader controller chosen so that the leader drives the swarm to the goal located at $\vect{p}_{G}$ while staying close to the average position of the swarm followers $\bar{\vect{p}}_{\alpha} = \frac{1}{N-1}\sum_{j\ne L}\vect{p}_j$.

\subsubsection{Interaction forces}

To model the interaction forces between the prober and the swarm agents, we employ a neighborhood-based approach. We define the  neighborhood of the prober as the set:
\begin{equation}
    \label{eq:prober_neigh}
  \neigh{p} \coloneqq  \{i \mid 0 < d_{ip} \le \overbar{r}_p, \forall i\in\{1,\dots, N\}\} 
\end{equation}
where $d_{ip} = \norm{\Delta \vect{p}_{ip}}$ is the $\ell_2$ Euclidean distance between the $i$\textsuperscript{th} swarm agent and the prober $p$, and $\overbar{r}_p$ is the sensing radius of the prober. The interaction forces are the same for all agents, including the unknown leader, and are calculated as:
\begin{align}
    \label{eq:force_model}
   \vect{f}_{pi}(t) &= [1_{i\in \neigh{p}}]\dfrac{\vect{p}_i-\vect{p}_p}{\norm{\diff{p}{i}{p}}^2}
\end{align}
where $[1_{i\in \neigh{p}}]=1$ if the $i$\textsuperscript{th} swarm agent is a neighbor of the prober and $[1_{i\in \neigh{p}}]=0$ if it is not.

\subsection{Observation Graph Features}

As we mentioned in the problem formulation section, even though the swarm graph evolves continuously with time, the prober observes it partially and in snapshot $o_k=\hat{\mathcal{G}}[k]=(\hat{\mathcal{H}}[k],\hat{\mathcal{E}}[k], k)$ at discrete time steps $k \in \mathbb{N}$.

\subsubsection{Temporal Features}
Although the prober does not have access to the actual time that has passed from the start of the episode, it can keep track of how many snapshots it has observed. We use these timestamps as a temporal feature in order to distinguish between states that differ in time but are identical otherwise.

\subsubsection{Node-level Features}
Every swarm member in the environment is a node in the graph. Member $i$ with, $i=1,\cdots N$ has a node feature representation encoded in the vector $\vect{h}_{i}\in \real{D_n}$. The prober is also part of the node set and has its own vector representation $\vect{h}_{p}\in \real{D_n}$. We define the node set, at time step $k$, as the set $\hat{\mathcal{H}}[k]\coloneqq \{\vect{h}_{1}[k],\cdots \vect{h}_{N}[k], \vect{h}_{p}[k]\}$.

The only \textit{node} feature used is the \textit{normalized relative position (NRP)} of each swarm agent measured from the prober's location. Feature normalization is a widely used technique in machine learning \cite{pmlr-v48-duan16}, and in the RL setup, it has been shown to accelerate training and improve generalization to unseen environments across a wide range of tasks \cite{andrychowicz2021what, huang2022cleanrl}.

\subsubsection{Edge-level Features}

As an episode evolves with time, the prober approaches the swarm and interacts with it. Keeping track of its interactions with each agent of the swarm, it calculates the \textit{interaction occurrences (IO)} feature. IO is an edge feature in the graph, and it measures the importance of each agent as a node in the graph for the prober based on the interactions it already has with it. The edges of the graph snapshot are partitioned into swarm-only (SO) and swarm-prober (SP) edges. From the perspective of the prober, the swarm part of the graph is assumed to be fully connected. Thus, the SO edges are bidirectional and always connected. That is, the edge feature vectors associated with the two edges $i\rightarrow j$ and $j\rightarrow i$ ($i,j=1,\ldots,N$) are:
\begin{equation}
    \vect{e}_{i\rightarrow j}[k] = \vect{e}_{j\rightarrow i}[k] = 1, \forall k
\end{equation}

The SP edges are unidirectional, with the prober always being the receiver and the swarm members the senders. The feature vectors associated with these edges are calculated based on the prober's interactions with the swarm agents as the episode evolves. Similar to \cite{bachoumas_icra_2025}, we count an interaction as valid if the distance of the prober to the $i$\textsuperscript{th} swarm agent is less that the same radius of interaction $\bar{r}_p$ that was used to define the neighborhood of the prober in \eqref{eq:prober_neigh}:

\begin{equation}
    \label{eq:counting}q^{i}[k] =
    \begin{cases}
        1 & \text{, if } \norm{\vect{p}_p[k]-\vect{p}_i[k]}\le \bar{r}_p  \\
        0 & \text{, otherwise}
    \end{cases}
\end{equation}

Using this definition, we aggregate interactions up to time step $k$ for the $i$\textsuperscript{th} agent as $q^{i}_{0:k}= \sum_{j=0}^{k}{q^i[j]}$ and define the edge features between the prober and the swarm agents as:

\begin{equation}
    \vect{e}_{i\rightarrow p}[k]=1+\frac{q^{i}_{0:k}}{\sum_{i=1}^{N}{q^i_{0:k}}},
    \forall~i=1,\dots,N
    \label{eq:quantified_interactions}
\end{equation}

We also stack all cumulative interactions in the interaction vector $\vect{q}_{0:k}= [q^{1}_{0:k}\dots,q^{N}_{0:k}]$.

\subsubsection{Graph-Level Features}

The graph-level features are used to capture the general motion and aggregate kinematic state of the swarm from the
perspective of the prober. Specifically, they quantify the collective translational and rotational dynamics of the swarm agents based on their positions at consecutive time-steps. The addition of these graph-level features was crucial for the success of our method. 

The key graph-level features are:

\paragraph{Mean Speed ($\bar{\vect{v}}$)} Measures the average instantaneous speed of individual agents.

\paragraph{Speed Variance ($\sigma^2_\vect{v}$)} Quantifies the dispersion in individual agent speeds around the mean speed.

\paragraph{Swarm Direction ($\hat{d}_{swarm}$)} Represents the normalized direction of the collective swarm movement, based on centroid displacement.

\paragraph{Swarm Centroid Speed ($\vect{v}_{swarm}$)} Measures the speed of the swarm's centroid.

\paragraph{Directional Alignment ($\bar{a}$)} Indicates how well, on average, individual agent movements align with the overall swarm direction.

\paragraph{Alignment Variance ($\sigma^2_a$)} Measures the variance in alignment values across agents, indicating the level of directional consensus. Low variance suggests high consensus (either aligned or anti-aligned depending on $\bar{a}$), while high variance indicates disordered movement relative to the swarm direction.

\paragraph{Swarm Angular Momentum ($\bar{L}$)} Computes the (scalar) average angular momentum of agents relative to the \emph{previous} swarm centroid $C_{prev}$.
This captures the net rotational motion around the swarm center.

\paragraph{Rotational Tendency ($\bar{R}$)} Measures the average tendency of agents to move tangentially with respect to the \emph{previous} swarm centroid $C_{prev}$. 

\subsection{Action space}

The action of the prober is two-dimensional and is defined as $a_k=\vect{v}_{p}[k]$ at time step $k$. The two components are the robot's base velocity in the local $x$-$y$ frame of the prober. For this work, the components of the base velocity of the prober are in the range $[-0.3,\dots,0.3] \frac{m}{sec}$, discretized with a step of
$0.05\frac{m}{sec}$. We found that this discretization achieved satisfactory results while more closely resembling a zero-order-hold (ZOH) implementation on a real robot compared to a continuous velocity profile.

\subsection{Reward function}

The reward function provides the guiding information necessary for training the probing policy. The reward scheme used to train the prober consists of 3 components discussed below.

\subsubsection{Maximize Leader Interaction (MLI) Term}
Since the primary objective is to identify the swarm's leader via an interactive scheme, the principal reward function is designed to maximize engagement between the prober and the leader agent. This reward is formally termed the Maximize Leader Interaction (MLI) reward. In \cite{bachoumas_icra_2025}, the authors used the unit softmax function to convert the interactions vector $\vect{q}_{0:k}$ into a probability distribution $Q_{k}$ at every iteration $k$. In the search for an expressive reward signal, we revisited the choice of the softmax function. Our analysis indicates that the unit softmax function's tendency to drive outputs toward the extremes of 0 or 1 is problematic during the initial stages of training. This behavior can prematurely commit the policy and hinder necessary exploration, which is critical when interactions with the swarm leader are inherently scarce. On the other hand, the ratio of interactions provides a much broader range of probabilities and could lead to unnecessarily slow convergence. For these reasons, we propose the usage of the average of the two metrics as a better metric of interaction.

First, the ratio of interactions between the prober and agent $i$ is calculated using the interactions from \eqref{eq:counting}
divided by the total number of interactions with the swarm defined
as:

\begin{equation}
    \label{eq:ratio}R_i[k]=\frac{q^{i}_{0:k}}{\sum_{j=1}^{N}{q^j_{0:k}}}
\end{equation}

By applying the unit softmax function to the interactions of the prober with the swarm, we get the second metric:

\begin{equation}
    \label{eq:softmax}S_{i}[k]=\frac{\exp(q^{i}_{0:k})}{\sum_{j=1}^{N}{\exp(q^j_{0:k})}}
\end{equation}

Finally, the average of the two base metrics is used as:

\begin{equation}
    \label{eq:mixture}Q_{i}[k]=\frac{R_{i}[k]+S_{i}[k]}{2}
\end{equation}

Here we introduce an additional modification to the MLI term from \cite{bachoumas_icra_2025}.
The authors did not take into consideration the following edge case. If the prober has gathered some interactions but then stops engaging with the swarm -- for instance, due to moving away -- it would still receive the same reward for the rest of the episode. This is not a desirable behavior since the prober should not be rewarded for losing contact with the swarm. To address this issue, we multiply the main reward by a mask $1_{d}[k]$ that is equal to $1$ if the distance between the prober and the leader did not increase between steps $k-1$ and $k$; otherwise, it
is equal to $\dfrac{1}{4}$. Finally, to calculate the reward, we query the distribution using the leader's index. The reward is given by the formula:

\begin{equation}
    r_{MLI}=1_{d}[k]\cdot (N\cdot Q_{L}[k]-1) \label{eq:mli_rew}
\end{equation}

Essentially, the prober receives a reward proportional to the probability it
assigns to the leader, with a maximum reward of $N-1$. Conversely, it incurs a penalty
when the probability $Q_{L}[k] \leq \frac{1}{N}$, with a maximum penalty of $-1$.
This reward structure incentivizes the prober to position itself near the leader
to maximize $Q_{L}[k]$, which leads to simultaneously minimizing $Q_{i}[k]$ for
all $i \in \{1,\dots,N\}$, $i \neq L$, since probabilities add up to $1$.

\subsubsection{Leader Distance (LD) Term}
This term is designed to penalize the prober for being far away from the leader. The reward is computed as:

\begin{equation}
    r_{LD}=\norm{\vect{p}_p[k]-\vect{p}_L[k]}_{2}\label{eq:ld_rew}
\end{equation}
where $\vect{p}_{p}$ and $\vect{p}_{L}$ are the positions of the prober and the leader, respectively. This auxiliary reward is used to shape the total reward for the prober in the early stages of training, where interactions with the swarm are not frequent and thus the main reward is sparse \cite{sutton1998reinforcement}. The maximum penalty for this term corresponds to the maximum possible distance between the prober and the leader. Since the episode terminates when the prober strays too far from the swarm, this termination threshold defines the upper bound of the penalty.

\subsubsection{Action Smoothing (AS) Term}
This term is designed to penalize the prober for making aggressive maneuvers. The reward is determined by:

\begin{equation}
    r_{AS}=\norm{\vect{v}_p[k-1]-\vect{v}_p[k]}_{2}\label{eq:as_rew}
\end{equation}
where $\vect{v}_{p}[k-1]$ and $\vect{v}_{p}[k]$ are the velocities of the prober
at time steps $k-1$ and $k$ respectively. The maximum penalty for this term is the
maximum change in velocity of the prober. One example is when the velocity $\vect{v}_{p}[k-1]=[
- 0.3,-0.3]$ changes to $\vect{v}_{p}[k]=[0.3,0.3]$, the maximum value is $0.85$.

Finally, the total reward is calculated as the sum of all individual terms with their corresponding weights:
\begin{equation}
    r_{total} = 2\cdot r_{MLI}-0.05\cdot r_{LD}-0.05\cdot r_{AS}
    \label{eq:reward_total}
\end{equation}

\begin{figure*}[ht]
    \centering
    \includegraphics[width=0.9\textwidth]{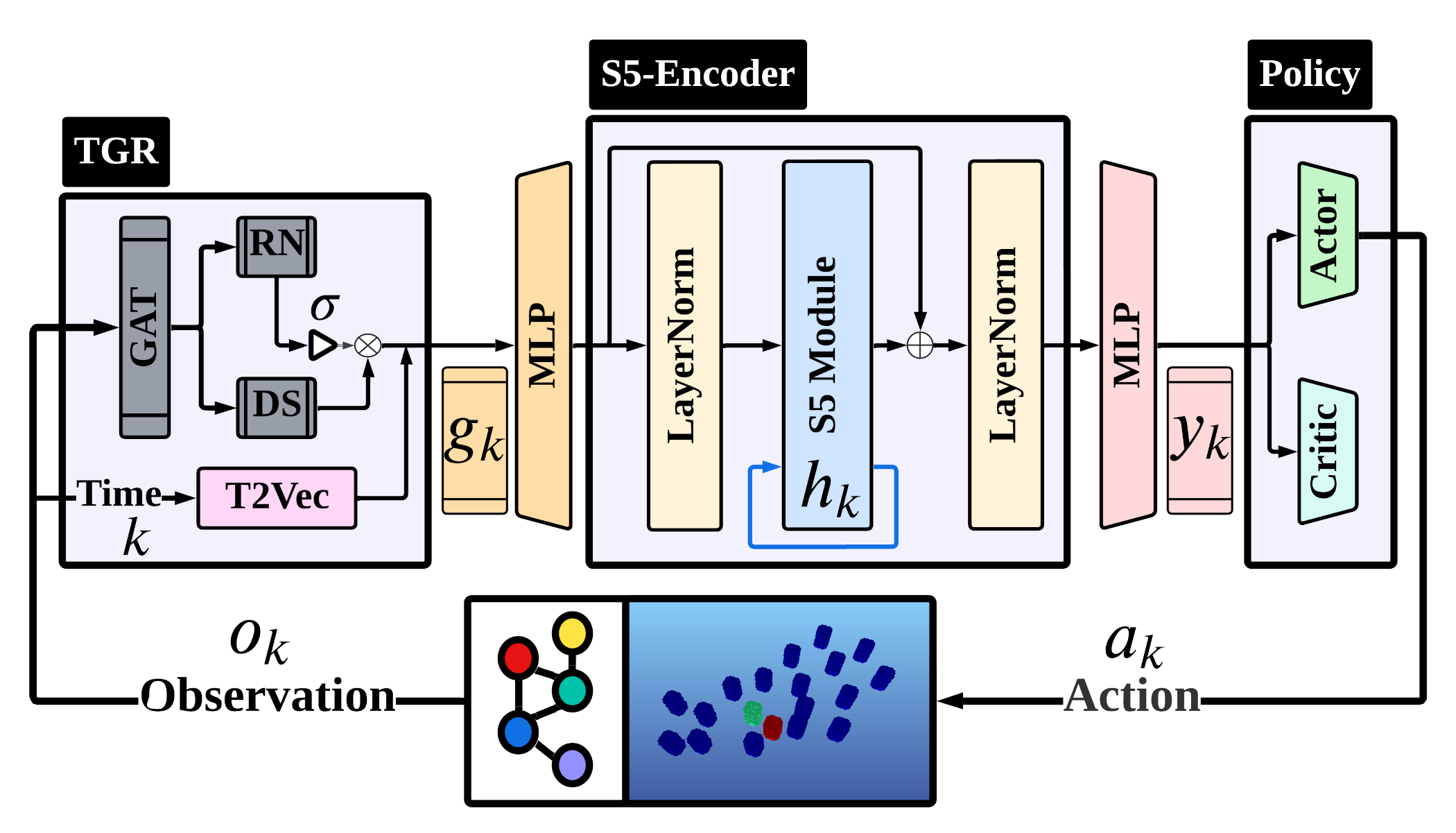}
        \vspace{-0.1in}
    \caption{Overview of the proposed \textit{prober} architecture. (a) At timestep $k$, the prober receives the graph snapshot observation $o_k=\hat{\mathcal{G}}[k]$ from the environment. The proposed TGR graph neural network layer processes the observation and produces a global graph representation $g_{k}$ by fusing the output of DeepSets (DS) and Relations Net (RN) using a gating mechanism. The S5 Encoder processes the output of the TGR layer and maintains internal state/context $h_{k}$ while producing an output encoding $y_{k}$. The encoding is fed to the Actor, which outputs an action $a_{k}$, and the Critic, which estimates the state value $v_{k}$. Finally, action $a_k$ is executed and the state of the environment changes.}
    \label{fig:s5_agent}
    \vspace{-0.1in}
\end{figure*}

%% file: policy.tex
\section{Probing Policy Design}

The proposed neural network architecture of the probing policy can be seen in Fig. \ref{fig:s5_agent}. We use the S5 architecture for efficient handling of the sequential nature of the iSLI task. Our key modification of the work in \cite{bachoumas_icra_2025} is the proposed graph neural network architecture, called Timed Graph Relationformer (TGR). The main task of the TGR layer is to create global graph representations of constant dimensionality that are permutation invariant with respect to the ordering of the node set and are used as input to the downstream S5 module. The TGR layer is designed to be able to capture the temporal dependencies between the graph snapshots, and to create informative global graph representations that are used for
the downstream policy learning task. The global graph representation
$\mathbf{g}_{TGR}[k]$ is projected to the appropriate dimensionality to be
processed by the S5-Encoder. Crucially, between each layer of the S5-Encoder, we perform pre-normalization and post-normalization using LayerNorm \cite{layer_norm} and use residual connections between each S5 layer. Finally, the output sequence $y_{t}$ is fed into a standard actor-critic module, which is composed of two distinct Multi-Layered Perceptrons (MLPs): an actor network to determine the subsequent action, and a critic network to evaluate the current state.

\subsection{Timed Graph Relationformer (TGR)}

The proposed TGR layer uses the following components:
\begin{itemize}
    \item Graph Attention Transformer (GAT) -  \cite{gat}

    \item Relations Net (RN) -  \cite{relationgnn}

    \item Deep-Sets (DS) -  \cite{deepsets}

    \item Time2vec (T2V) -  \cite{time2vec}
\end{itemize}

The forward pass of the TGR layer is as follows:

\begin{align}
    \hat{H'}[k]         & = \text{GAT}(\hat{H}[k], \hat{S}[k], \hat{R}[k]) \\
    \mathbf{g}_{DS}[k]  & = \text{DS}(\hat{H'}[k])                               \\
    \mathbf{g}_{RN}[k]  & = \text{RN}(\hat{H'}[k], \hat{S}[k], \hat{R}[k])       \\
    \mathbf{g}_{GR}[k] & = \mathbf{g}_{DS}[k] \odot \sigma(\mathbf{g}_{RN}[k])  \\
    \mathbf{g}_{TGR}[k] & = \mathbf{g}_{GR}[k] \oplus \text{T2V}(k)
\end{align}

The input to our proposed graph neural network is a graph snapshot
$\hat{\mathcal{G}}[k]=(\hat{H}[k],\hat{S}[k],\hat{R}[k], k)$. A Multi-head Graph Attention Transformer layer processes the node set of the partial observation graph $\hat{\mathcal{G}}[k]$ to create new node representations $\hat{H'}[k]$. Then these nodes are fed in parallel to a DeepSet (DS) layer and a Relation Net (RN) layer to produce global graph representations $\mathbf{g}_{DS}[k]$ and $\mathbf{g}_{RN}[k]$ separately. Then a gating mechanism is created by applying a sigmoid activation $\sigma(x)=\frac{1}{1+\exp(-x)}$ to the output of the RN layer. Then element-wise multiplication is performed between the gate and the output from the DeepSets (DS) layer. Finally, a Time2Vec module embeds the raw current timestep index $k$ and projects it into a vector of the same dimension as the global representation generated by the DS and RN gating mechanism. We concatenate the two vectors together to create the \textit{timed global graph representations} $\mathbf{g}_{TGR}[k]$ that are then passed into the S5 module and are processed as an input sequence.

The global graph representations are directly influenced by the node embeddings. Essentially, the better the network becomes at creating useful node representations, the more informative the global representations are going to be for the downstream policy learning task. This hypothesis is backed up by the fact that the model is trained end-to-end. We note here that different representations produced by the graph neural network can be used for different tasks at the node, edge, or graph level. We leave the possibility of using the node and edge embeddings for future work.


\subsection{Final leader identification task}

We formulate the final task of leader identification as a Bayesian estimation problem. In this framework, the leader's identity is treated as a random variable, and a Bayesian estimator is employed to update the belief about this identity using information gathered from the prober's interactions.

Formally, we define the leader's identity at timestep $k$ as a random variable, $L_k$. Assuming no prior information, we assign a uniform probability distribution over all agents at the initial timestep, $k=0$. Therefore, for a swarm of $N$ agents, the prior probability, $p(L_0)$, for any given agent being the leader is $1/N$. Then, we use the vector of accumulated interactions between the prober and the swarm up to time $k$, denoted by $\vect{q}_{0:k}= [q^{1}_{0:k}, \dots, q^{N}_{0:k}]$, as evidence to update the prior into the posterior probability, $p(L_k|\vect{q}_{0:k})$, via Bayes' rule:

\begin{equation}
    p(L_k|\vect{q}_{0:k}) = \dfrac{p(\vect{q}_{0:k}|L_k)p(L_k)}{p(\vect{q}_{0:k})}
\end{equation}

The term $p(\vect{q}_{0:k}|L_k)$ represents the likelihood of observing the interaction history $\vect{q}_{0:k}$ under the assumption that the leader's identity, $L_k$, is known. To approximate this likelihood, we adopt a data-driven approach. First, a dataset is compiled by deploying a trained prober across a series of simulated test environments with varying swarm sizes and known leaders. For each simulation, we compute the proportion of interactions between the prober and each swarm member at every timestep. We then employ Gaussian Kernel Density Estimation (GKDE) \cite{hardle2004nonparametric} on this dataset to learn a non-parametric model of the likelihood function. This process yields a robust approximation of $p(\vect{q}_{0:k}|L_k)$ derived directly from empirical data.

With the likelihood model defined, we implement a recursive Bayesian update scheme, where the posterior from the previous timestep serves as the prior for the current one. This estimation process is robust to erroneous intermediate beliefs due to the efficacy of the offline prober training. This stands in contrast to the method presented in \cite{bachoumas_icra_2025}, which lacked a formal uncertainty quantification mechanism and relied solely on the prober's proximity to swarm agents as a heuristic for leadership identification.

\subsection{Parameter grid-search} 

We designed an extensive procedure to find the best parameters for our model. We performed a grid search over the architecture-related hyper-parameters, the PPO parameters, and the environment parameters. The models were all trained for $100$M time-steps, and except for the tested parameters, all other parameters were kept the same. All parameters are listed in Table \ref{tab:gridsearch}.

\begin{table}[t!]
    \caption{Model parameters in the grid search.}
    \centering
    \setlength{\tabcolsep}{5pt}
    \begin{tabular}{|c||c|}
        \hline
        \textbf{Term}             & \textbf{Range} \\
        \hline
        \hline
        PPO Clip Ratio :          & $[0.2,0.3]$    \\
        \hline
        PPO Entropy Coefficient : & $[0.01,0.02]$  \\
        \hline
        S5 model dimension        & $[128, 256]$   \\
        \hline
        S5 number of layers       & $[2, 4]$       \\
        \hline
    \end{tabular}
    \vspace{5pt}
    \label{tab:gridsearch}
    \vspace{-8pt}
\end{table}

From the grid-search analysis, the most effective combination is with parameters $0.2$ for the PPO clip ratio, $0.01$ entropy coefficient, and a model size of $256$ with $4$ layers for the S5 encoder. Training one policy for $100M$ steps took approximately one hour, and we performed a total of $24$ runs.

%% file: sim_experiments.tex
\section{Model Evaluation - Simulation}

\subsection{Comparison with Baseline Models}
We performed extensive generalization experiments for the prober's policy and compared the effectiveness of our proposed graph neural network layer against the
following baseline models:

\begin{itemize}
    \item \textbf{DS}: A DeepSets \cite{deepsets} model.
    \item \textbf{GAT+DS}: A Graph Attention Transformer \cite{gat} and DeepSets combined model.
    \item \textbf{GAT+DS+RN}: A Graph Attention Transformer, DeepSets, and a Relations Net combined model \cite{relationgnn}.
\end{itemize}

The first baseline model uses a DeepSets (DS) layer. The DS layer views the
graph as a set and thus ignores all connectivity information. It first passes the
node representations via a learned nonlinear transformation, typically an MLP, concatenates
the result with the input global graph representations, and feeds them to another
learnable nonlinear function, again an MLP. Since we encode the interactions as
edge features in the iSLI problem, we expect this model to perform poorly.

The second baseline model uses a Graph Attention Transformer (GAT) + DS
architecture. The GAT layer processes the input graph and fuses the node representations
with an attention mechanism that respects the topology of the graph by using information
from the sender and receiver sets, thus taking edges into consideration. A GAT layer
ignores the global graph features, so we pair it with a DS layer to construct a more
informative graph representation.

The third baseline model adds a Relations Net \cite{relationgnn} layer to the aforementioned
GAT + DS architecture to create the GAT+(DS+RN) model. Similar to the second model,
the GAT layer creates better node representations using attention. Then it feeds
these nodes in parallel to the DS and RN layers that create two separate global graph
representations that we can add or concatenate. We choose to add the two graph
representations.

Therefore, for this evaluation, the only difference between the models is the graph neural network used before the S5 encoder. The environment is set to have $N = 15$ swarm
agents that travel with a maximum speed of $v_{max}= 0.3 \frac{m}{s}$. The purpose of this experiment was to compare the proposed architecture against the aforementioned baselines and determine the margin of performance it can achieve.

\subsection{Generalization Tests}

We use the trained model employing our proposed graph neural network architecture
and test its performance in environments that are out of the training distribution.
We note that we did not fine-tune the model to achieve any of these results, as we
are interested in its zero-shot performance.

Moreover, we tested the model's generalization capabilities when the number of swarm agents differs from training and when the swarm moves at different speeds. Both changes result in observation sequences that differ significantly from those seen during training. This test also evaluates whether the prober has truly learned the task or has simply overfitted to the swarm's speed and is exploiting this to maximize its rewards.

%% file: sim_results.tex
\section{Simulation Results}

All simulations for testing the proposed architecture were conducted on a computer equipped with a 12th Gen Intel\textsuperscript{{\textregistered}}
Core\textsuperscript{\texttrademark} i9-12900K processor featuring 24 cores, as
well as an NVIDIA\textsuperscript{{\textregistered}} RTX A5500 GPU, running Linux
Ubuntu 22.04 LTS. The framework is built using Python 3.10.12 and JAX, a library
for high-performance scientific computing \cite{jax}.

\subsection{Training Performance Comparison}

\begin{figure}[t!]
    \centering
    \includegraphics[width=1\columnwidth]{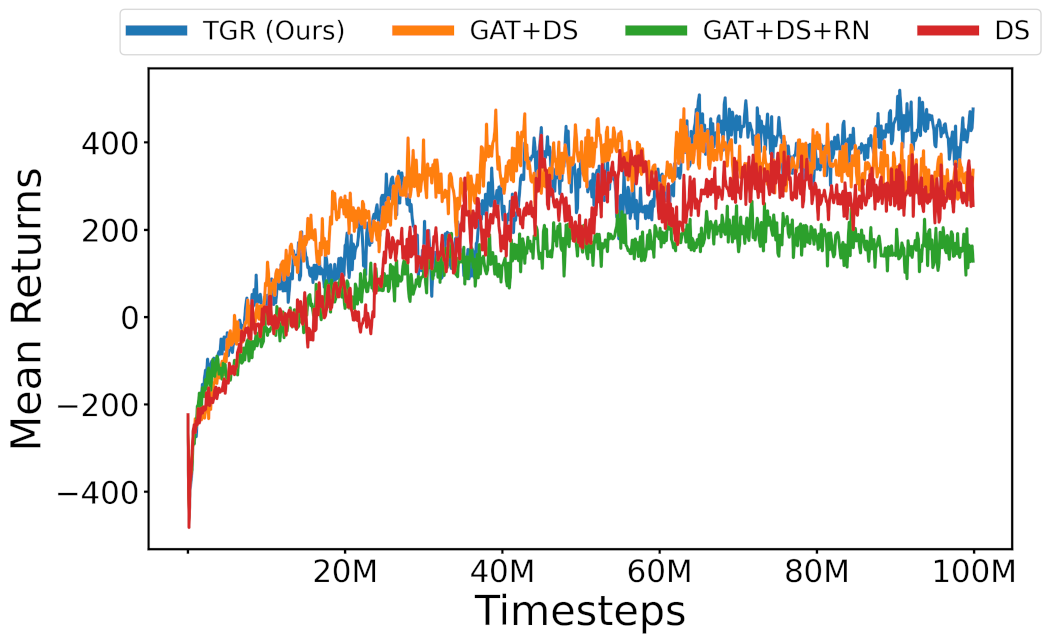}
     \vspace{-0.2in}
    \caption{Mean returns for different graph neural network backbones. Our proposed
    TGR architecture outperformed all baselines, shown by the highest returns at the end of training. Results are averaged over 5 runs for 100M timesteps of training using different randomization seeds.}
    \label{fig:mean_return_plot}
        \vspace{-0.2in}
\end{figure}

Figure \ref{fig:mean_return_plot} shows the accumulated returns for different graph neural network backbones (baseline models), while the rest of the network and the environments remain the same. Due to JAX's controlled pseudo-randomness \cite{jax}, all randomization seeds used for the environments and PPO are guaranteed to be the same for all networks, thus offering an even playing field that makes comparative analysis reliable.

We can see that our proposed TGR model outperformed all baseline models during training, achieving the highest mean returns. Interestingly, the model that performed the worst was not the simplest one, i.e., the DS model, but the GAT+(DS+RN) model. While GAT+(DS+RN) shares the same components as our proposed TGR architecture, its under-performance suggests that the method used to combine these different information streams is a critical design choice.

Specifically, the GAT+(DS+RN) model processes node representations with GAT and then feeds these into DS and RN layers in parallel, combining their outputs through simple addition to form the final graph representation. Naively adding the output of the relational component to the set-based component fails to be effective because the values from these two components have different scales. By analyzing the individual outputs of each layer, we found that the components of the output vector from the RN layer are $1 \sim 2$ orders of magnitude smaller than those from the DS layer. Thus, adding them together essentially leads to discarding the set-based information coming from the RN layer.

The relations-gating mechanism employed by our proposed TGR model modulates the output of the DS layer through element-wise multiplication. This learned gating allows relational insights to dynamically influence the set-based representation, providing a more nuanced and potentially more effective way to fuse these information types compared to simple addition.

\subsection{Generalization Test}

\begin{figure}[t!]
    \centering
    \includegraphics[width=1\columnwidth]{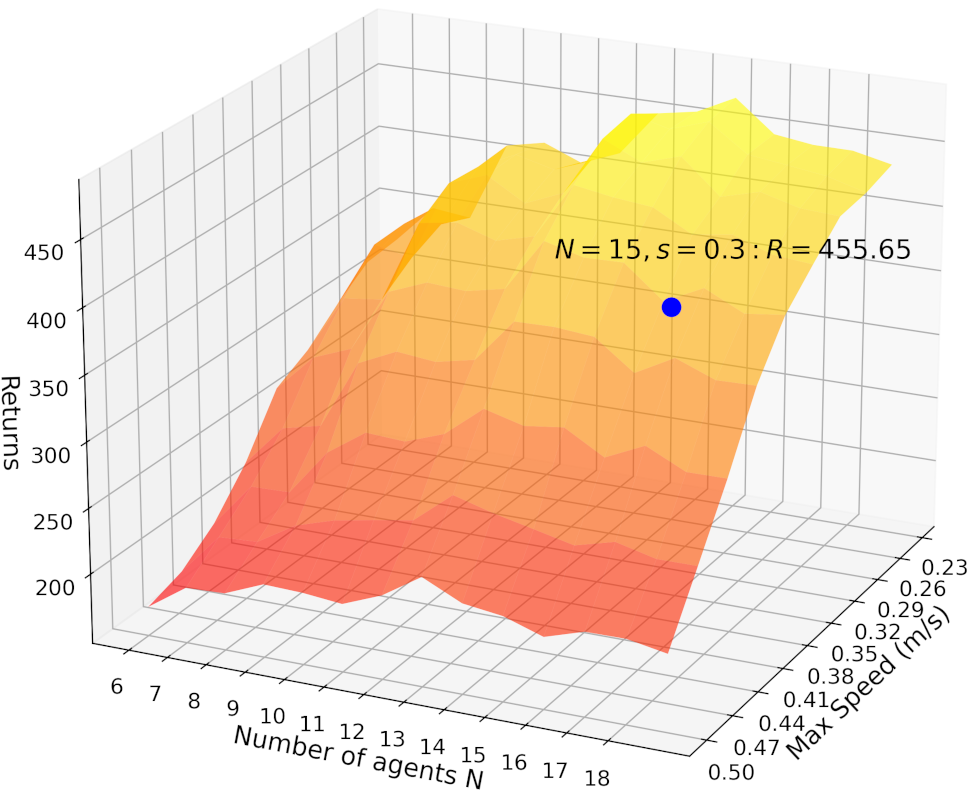}
    \vspace{-0.3in}
    \caption{Performance, measured by returns, of the proposed TGR model trained in an environment with $N = 15$ agents and maximum speed $v_{max} = 0.3\, \frac{m}{sec}$ (blue dot), evaluated across environments with varying values of $N$ and $v_{max}$. Each point on the surface represents the average return over $5$ random seeds and $1000$ environments.}
    \label{fig:speed_num_agents}
\end{figure}

\begin{figure}[t!]
    \centering
    \vspace{-0.2in}
    \includegraphics[width=1\columnwidth]{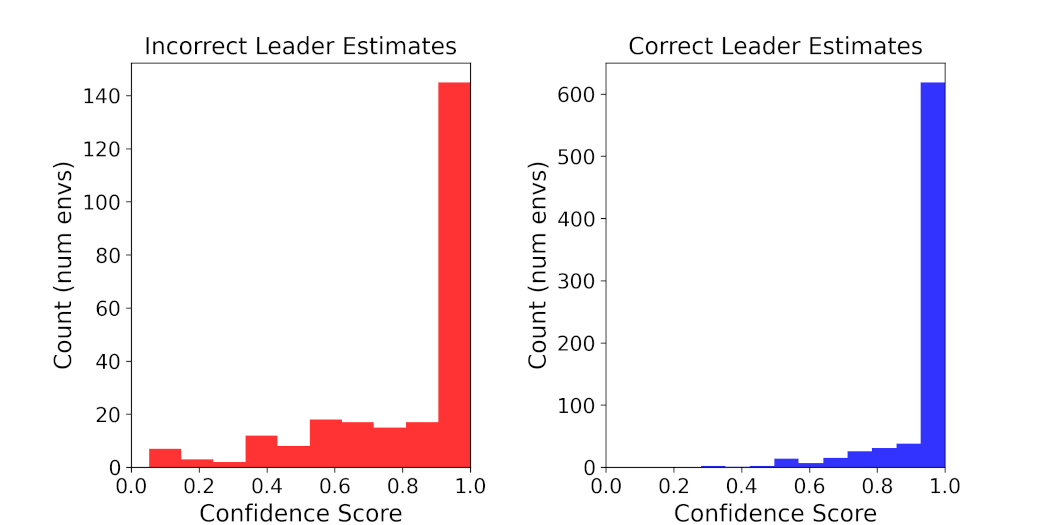}
    \vspace{-0.2in}
    \caption{Leader identification as a confidence score for an environment that is not from
    the training distribution, with $N=19$. On the left, the red histogram shows the distribution of confidence scores assigned to wrong leader predictions. On the right, the blue histogram shows the distribution of confidence scores in correct leader predictions. The prober is strongly confident in its correct decisions in $758$ of the $1000$ environments or in $75.8\%$ of them.}
    \label{fig:confidence_test}
    \vspace{-0.2in}
\end{figure}
\begin{figure}[t]
    \centering
    \includegraphics[width=1\columnwidth]{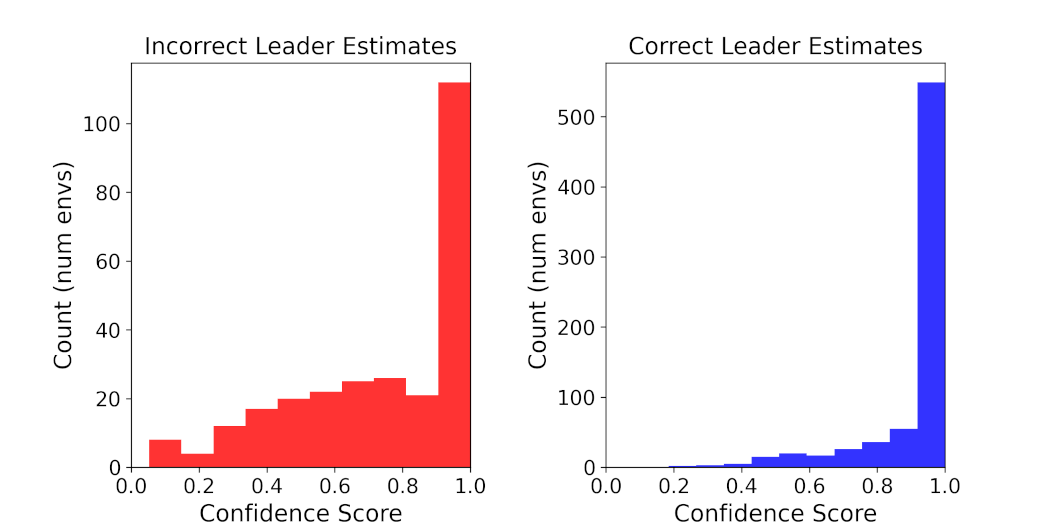}
    \vspace{-0.2in}
    \caption{Leader identification as a confidence score for an out-of-distribution environment with $N=19$ and $v_{max}=0.5 \frac{m}{sec}$. On the left, the confidence in wrong leader
    predictions, and on the right, the confidence in correct leader predictions.
    The prober is correct in $707$ out of $1000$ environments or $70.7\%$ of them.}
    \label{fig:confidence_test_speed}
        \vspace{-0.2in}
\end{figure}

We use the TGR model that performed best in the grid-search of the previous section and train
it for $100$M time-steps in the $N=15$ environment. After training, we evaluated
the model across $1000$ out-of-distribution environments for varying numbers of
agents in the range $N = [6,\dots, 19]$ and different maximum speeds for the agents
$v_{max}= [0.23,\dots, 0.5]\frac{m}{s}$. In Fig. \ref{fig:speed_num_agents}, we can see a surface plot that shows the performance of the trained model across all different environments. We also note the single point of training with a blue dot. All other points were calculated as the zero-shot performance without finetuning.

As shown, the model demonstrates significant generalization performance, achieving high returns across a wide range of tested parameters (number of agents $N$ and maximum speed) that differ from those used during training. Here, we want to note that the varying speed only refers to the speed of the swarm agents; the prober is limited to a speed of $0.3 \frac{m}{sec}$ as it was specified in the environment section.

This result demonstrates the strong capabilities of the proposed architecture in zero-shot performance to unseen environments for the iSLI task. Furthermore, it provides a reliable indication that this performance can be transferred from simulation to reality without further training.

\subsection{Leader identification}

We evaluated the performance of the prober in the leader identification task by an additional measure called the \textit{confidence score}. Using the Bayesian estimator we described in the previous section, we define the confidence of the prober in estimating the identity of the leader as the last probability that was assigned to the estimated leader. A well-calibrated prober should exhibit confidence that is indicative of its accuracy. Specifically, it should report a high confidence score when correctly identifying the leader and, conversely, a low confidence score when its estimate is erroneous.

We conducted two evaluations using the model trained with $N = 15$ swarm agents and $v_{max} = 0.3\, \frac{m}{sec}$. The model was tested on $1000$ out-of-distribution environments with $N = 19$ and $v_{max} = 0.3\, \frac{m}{sec}$, and another $1000$ out-of-distribution environments with $N = 19$ and $v_{max} = 0.5\, \frac{m}{sec}$.

As Fig. \ref{fig:confidence_test} demonstrates, the prober is able to maintain a high level of performance in out-of-training distribution environments. Specifically tested in a more challenging scenario with an increased swarm size ($N=19$) and a maximum speed of $v_{\text{max}}=0.3\,\text{m/s}$, the prober correctly identified the leader in $75.8\%$ of the 1,000 trials. A comparison of the confidence distributions reveals that the prober is well-calibrated: in successful identifications, it exhibits high confidence, whereas in the $24.2\%$ of mistaken cases, it frequently reports low confidence in its incorrect estimate.

Finally, Fig.~\ref{fig:confidence_test_speed} presents the policy's performance on the most demanding generalization task, evaluated over 1,000 environments with an increased swarm size ($N=19$) and speed ($v_{\text{max}}=0.5\,\text{m/s}$). In this challenging setting, the prober correctly identified the leader in $70.7\%$ of trials. Consistent with previous findings, in the remaining $29.3\%$ of incorrect identifications, the prober again demonstrated greater uncertainty in its estimates.

%% file: real_experiments.tex
\section{Robot Experiments}

\begin{figure*}[t!]
    \centering
    \includegraphics[width=\textwidth]{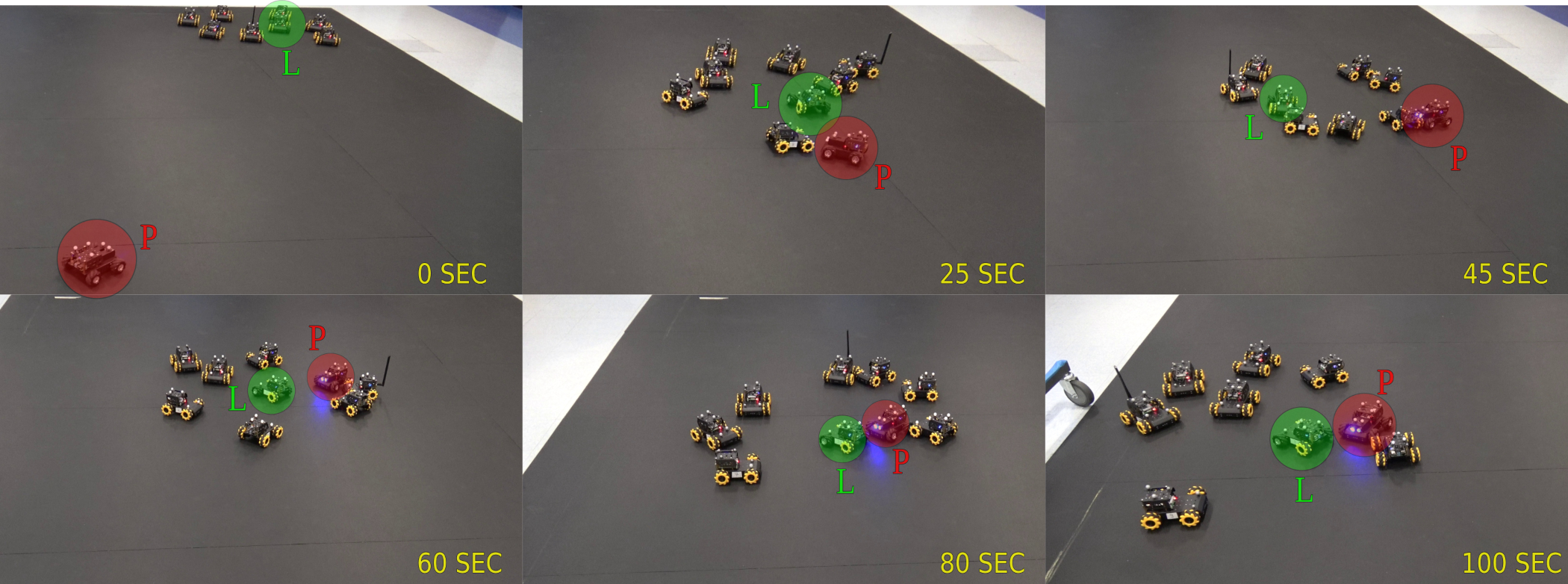}
    \caption{Six stages of a real-world experiment where the prober (red circle) interacts with a maneuvering swarm to identify its unknown leader (green circle). The sequence progresses from top left to bottom right, showing the prober actively engaging with the swarm as it moves toward its destination.}
     \vspace{-0.25in}
    \label{fig:robot_exp_setup}
\end{figure*}

To further validate our approach, we performed a series of real-world robot experiments. The experimental setup employed eight Hiwonder TurboPi \cite{hiwonder} robots equipped with mecanum wheels. The control and planning of the robots is handled using ROS2 \cite{ros2}. Seven of these robots formed the swarm, while one acted as the prober, as seen in Fig.~\ref{fig:robot_exp_setup} (top row). The arena the robots were moving in was $16'\times14'$ in size and covered with rubber floor mats. We accurately track the robots' positions at $120\,$Hz using an Optitrack Motion Capture System comprising eight cameras: six Primex 13 and two Primex 13W cameras (OptiTrack - Motion Capture Systems). This setup enabled us to transition from simulation to a physical environment and test our algorithm's performance under real-world conditions. Here we present two sets of experiments.

\subsection{Generalization to Different Swarm Characteristics}

The first set of experiments demonstrates the generalization capabilities of a trained prober. We used a model that was trained with $N=15$ agents in the swarm and deployed it to the aforementioned testbed with $N=7$ swarm agents. Apart from the different sizes of the swarm, other key differences between the real testbed and the simulated environments make the transition very challenging. For instance, although the employed robots are holonomic systems, there are still unmodeled nonlinear effects, such as wheel slipping and skidding, as well as motor saturation. These differences lead to completely unseen observation trajectories from the prober's perspective, as the flocking patterns arising from the swarm are much more complex. Moreover, in the real testbed, relevant events occur at different frequencies. Specifically, the trained policy runs at $5\,Hz$ on the Raspberry Pi microcontroller onboard the prober robot, whereas in simulation it was executed at a faster rate of $20\,Hz$. All of the above discrepancies between simulation and reality, collectively referred to as the \textit{sim-to-real gap}, challenge the generalization capabilities of the proposed methodology to their limits.

Figure~\ref{fig:robot_exp_setup} illustrates a representative experiment, with the prober highlighted by a red circle and the leader by a green circle. The sequence begins (top left, 0 s) with the prober positioned far from the swarm. As it approaches, it first interacts with a follower agent (25 s) before maneuvering into the center of the swarm to engage with other members. By the end of the trial (bottom right, 100 s), the prober has focused its interactions and successfully identified the leader, demonstrating an effective identification-through-interaction strategy.

\begin{figure}[t!] 
    \centering   \includegraphics[width=1\columnwidth]{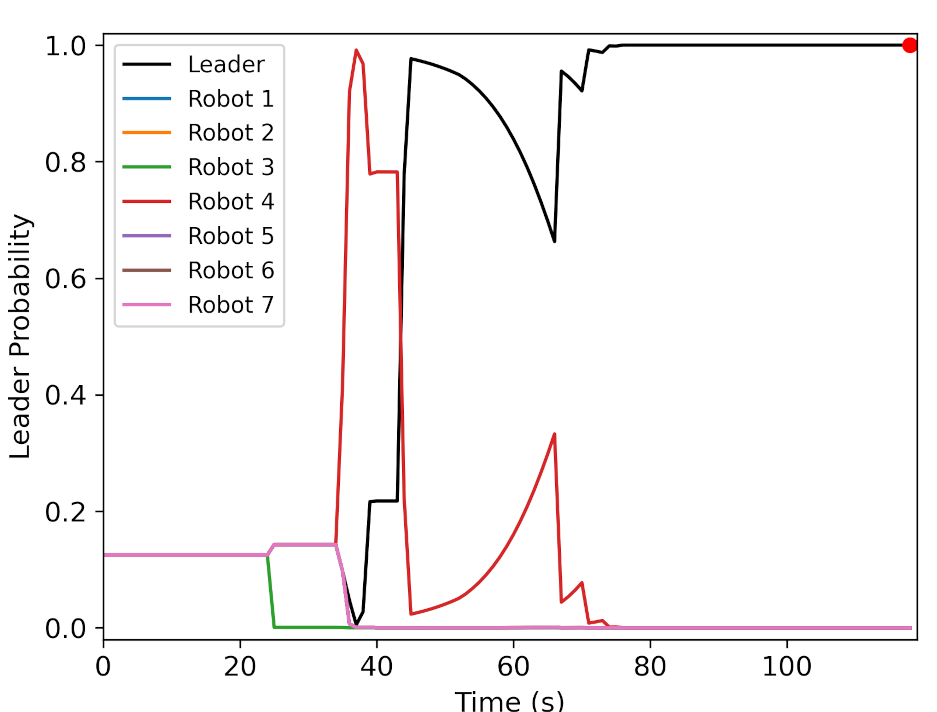}
    \vspace{-0.2in} 
    \caption{This figure depicts the leader's estimated probability in a generalization experiment with a swarm of $N=8$ robots, despite the policy being trained in simulation with $N=15$ agents. The prober initially interacts with the entire swarm, successfully leveraging this interaction data to correctly infer the leader's identity.}
    \label{fig:generalization_exp1} 
    \vspace{-0.2in}
\end{figure}

Figure~\ref{fig:generalization_exp1} demonstrates the zero-shot generalization performance of the policy in a real-world experiment with a swarm of $N=8$ robots. Although trained exclusively in simulation with a larger swarm of $N=15$ agents, the prober rapidly identifies the correct leader and maintains high confidence throughout the episode. This result validates the successful sim-to-real transfer of the policy and highlights its robustness to changes in swarm size and complex, real-world dynamics.

\subsection{Robustness to Unexpected Observation Change}

The second set of experiments demonstrates the prober's capability to adapt to a sudden change in the number of agents in the swarm. The same model as in the previous experiment is used. In this scenario, we evaluate the prober's robustness to unforeseen agent disconnections. The experiment commences with a swarm of seven agents. During the task, a simulated fault is introduced that causes the prober to permanently lose sensory contact with two of the follower agents. Despite this partial loss of information, the prober must still complete the iSLI task under these degraded conditions. This experiment tests the robustness of the prober to sudden and unexpected changes in its observation input.

\begin{figure}[t!] 
    \centering
    \includegraphics[width=1\columnwidth]{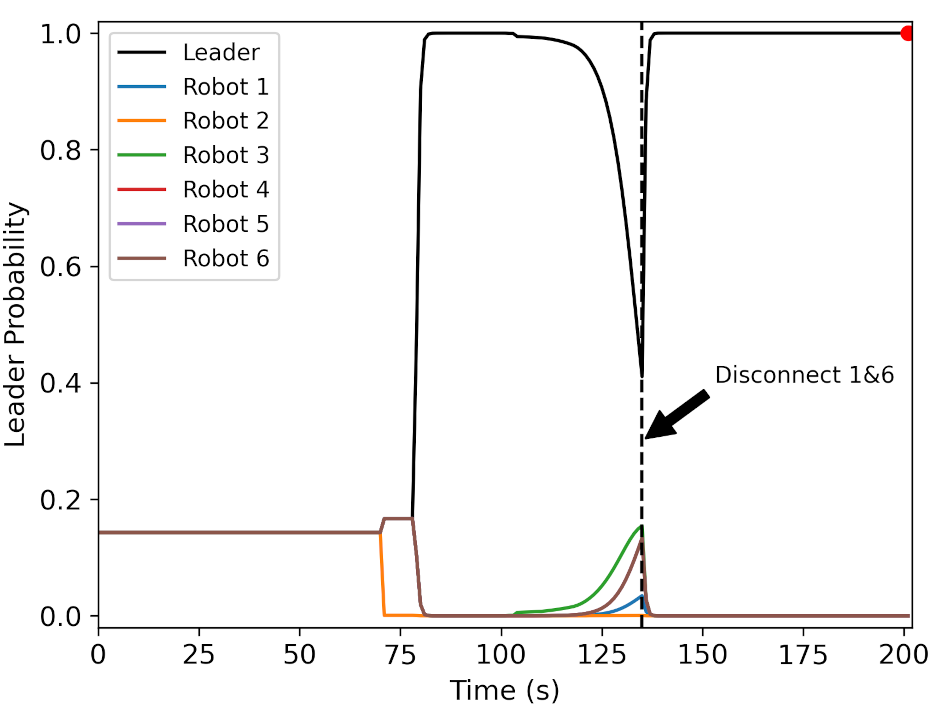}
    \vspace{-0.2in} 
    \caption{Demonstration of the policy's robustness to an abrupt agent disconnection event. Mid-mission, two follower agents (Agents 1 and 6) are permanently removed from the prober's observation space. The plot shows the prober adapting to this sudden and unforeseen change in real-time. After correctly identifying the leader (Robot 0, black line) prior to disconnection, it maintains a significant belief ($p_L=40.7\%$) during the event and subsequently re-converges on the correct identification.}
    \label{fig:disconnect_exp}
\end{figure}

Figure~\ref{fig:disconnect_exp} illustrates the policy's robustness to a sudden loss of sensory information. In this experiment, two follower agents (Agents 1 and 6) were disconnected from the prober's view 131 seconds after the experiment started. The disconnection event occurred when the correct leader's probability was high (93\%) and caused a temporary drop to 40.7\%. However, the estimator rapidly recovers by leveraging new interactions, and the final probability redistribution further confirms the correct leader. This experiment highlights the methodology's ability to adapt to significant, unforeseen changes in swarm observability.

Collectively, these real-world experiments validate our simulation findings, demonstrating that the proposed TGR-S5 architecture and RL training methodology yield policies that exhibit effective zero-shot sim-to-real transfer, robust generalization across different swarm configurations, and resilience to real-world complexities, including dynamic events such as sensor loss.

%% file: conclusions.tex
\section{Conclusions and Future Work}

This work presented the Timed Graph Relationformer (TGR) graph neural network layer and used it to effectively tackle the interactive Swarm Leader Identification (iSLI) problem. Our proposed layer, integrated with a Structured State Space Sequence (S5) model, demonstrated high efficacy in processing the sequential, graph-based observations inherent to the iSLI task. A key factor in our method's success was the architecture's learned gating mechanism, which skillfully fused relational and set-based graph information, outperforming simpler baseline models. The trained probing policy exhibited impressive zero-shot generalization, performing reliably in simulated environments with diverse swarm sizes and speeds beyond its training parameters, without needing further fine-tuning. The prober consistently achieved high accuracy in identifying the leader across various scenarios, both within and outside the training distribution, and its confidence scores accurately reflected the certainty in its prediction. Crucially, the methodology proved transferable from simulation to real-world robotics, successfully identifying the leader in a physical swarm despite significant sim-to-real gaps like unmodeled dynamics and differing operational parameters. The system also displayed robustness against sudden environmental changes, such as the unexpected loss of sensor data from swarm agents, adapting effectively to maintain performance in this challenging and previously unseen scenario. 

Overall, this work provides a robust, generalizable, and experimentally validated framework for identifying hidden leaders within swarms through adversarial interaction. This intelligent prober represents an ideal tool for compelling a swarm to evolve more sophisticated and decentralized behaviors. This process forges a resilience with significant implications across a variety of applications, ranging from robust robotic swarm deployments to a deeper understanding of biological swarms and complex social interactions.

Future work could explore scaling up the proposed methodology to teams of multiple probers using Multi-Agent Reinforcement Learning (MARL) \cite {wang2022modelbasedmultiagentreinforcementlearning} that could identify potentially multiple leaders in much bigger swarms. Another promising direction for future work is reducing the computation time of the deployed policy. While the robots used in this study were slow ground vehicles, faster platforms such as drones would require the policy to run at significantly higher rates than the $5\,Hz$ achieved here. Finally, replacing the external motion capture system with onboard estimation methods, such as Visual-Inertial Odometry (VIO), represents another potential extension of this work for real hardware experiments.